\documentclass[sigconf, nonacm]{acmart}

\usepackage{mathtools}
\usepackage{amsmath}

\usepackage{amssymb}
\usepackage{amsthm}
\usepackage{color}
\usepackage{bm}
\usepackage{booktabs}
\usepackage{multirow}
\usepackage{arydshln}
\usepackage{enumitem}
\usepackage{dsfont}
\usepackage{bbding}
\usepackage{color, colortbl}
\usepackage{multirow}
\usepackage{booktabs}
\usepackage{caption}
\usepackage{colortbl}
\usepackage{xcolor}
\usepackage{multirow}
\usepackage{pifont}
\usepackage{graphicx}
\usepackage{threeparttable}

\newcommand{\supp}[1]{\emph{Supp.}}

\definecolor{Blue}{rgb}{0,0,1} 

\begin{document}

\title{\textsc{Phantom-Insight}: Adaptive Multi-cue Fusion for Video Camouflaged Object Detection with Multimodal LLM}

\author{Hua Zhang$^{1,2,*}$, Changjiang Luo$^{1,2,*}$}
\affiliation{$^{1}$ Institute of Information Engineering, Chinese Academy of Sciences \country{}}
\affiliation{$^{2}$ School of Cyber Security, University of Chinese Academy of Sciences \country{}}
\affiliation{\texttt{zhanghua@iie.ac.cn} \quad\quad \texttt{luochangjiang@iie.ac.cn}\country{}}
\affiliation{$*$Equal contribution\country{}}

\begin{abstract}
    Video camouflaged object detection (VCOD) is challenging due to dynamic environments. Existing methods face two main issues: (1) SAM-based methods struggle to separate camouflaged object edges due to model freezing, and (2) MLLM-based methods suffer from poor object separability as large language models merge foreground and background. To address these issues, we propose a novel VCOD method based on SAM and MLLM, called \textsc{\textbf{Phantom-Insight}}. \textit{To enhance the separability of object edge details}, we represent video sequences with temporal and spatial clues and perform feature fusion via LLM to increase information density. Next, multiple cues are generated through the dynamic foreground visual token scoring module and the prompt network to adaptively guide and fine-tune the SAM model, enabling it to adapt to subtle textures. \textit{To enhance the separability of objects and background}, we propose a decoupled foreground-background learning strategy. By generating foreground and background cues separately and performing decoupled training, the visual token can effectively integrate foreground and background information independently, enabling SAM to more accurately segment camouflaged objects in the video. Experiments on the MoCA-Mask dataset show that \textsc{\textbf{Phantom-Insight}} achieves state-of-the-art performance across various metrics. Additionally, its ability to detect unseen camouflaged objects on the CAD2016 dataset highlights its strong generalization ability.
\end{abstract}




\maketitle

\begin{figure*}[t] 
    \centering
    \includegraphics[width=0.8 \textwidth,keepaspectratio]{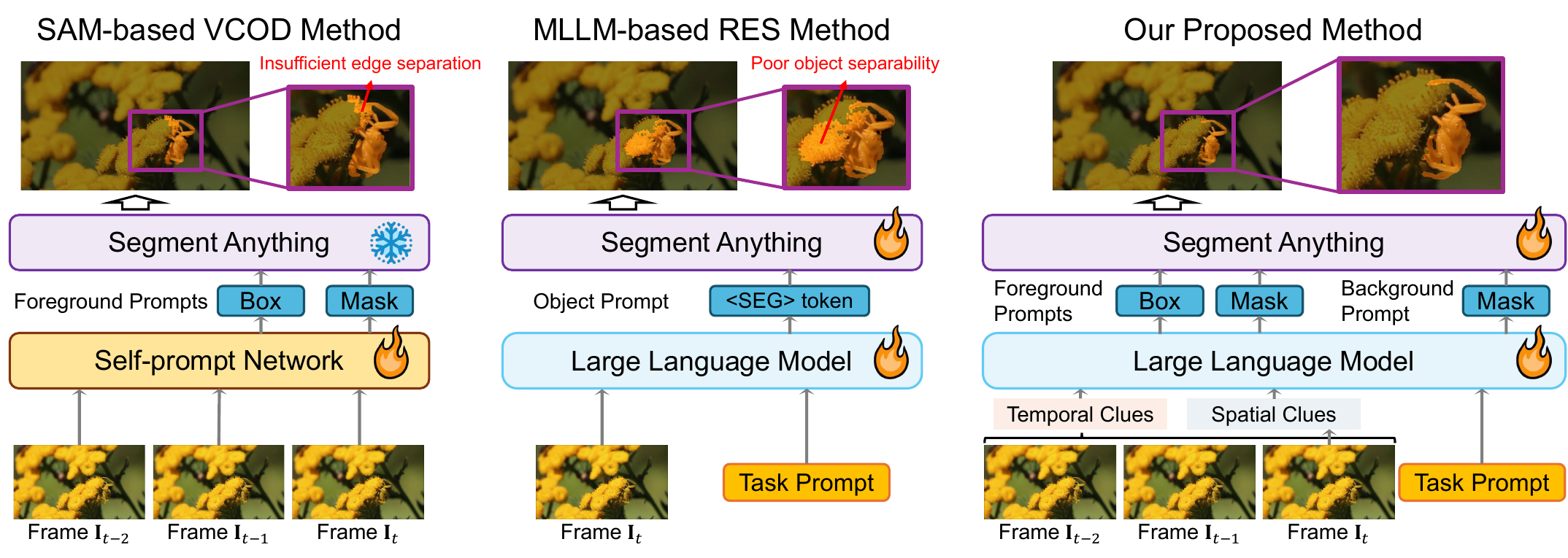} 
    \caption{Illustration of the previous methods and the proposed \textsc{\textbf{Phantom-Insight}} method. SAM-based VCOD method faces challenges with insufficient edge separation due to the frozen SAM parameters, while MLLM-based methods struggle with poor object separability as the LLM merges foreground and background information. Our method integrates temporal-spatial clues and LLM-enhanced feature fusion to generate multiple cues that refine object edge separation while employing a decoupled foreground-background learning strategy to ensure precise object segmentation.}
    \label{abstract}
\end{figure*}

\section{Introduction}

The development of computer vision technology has significantly advanced object detection techniques~\cite{zou2023object}, with one of the key challenges being the detection of camouflaged objects, which are often difficult for humans to identify. Camouflaged object detection aims to identify objects that blend with their background~\cite{tang2024chain,yu2024exploring,he2024text,niu2024minet,fu2024semi,huang2023feature, YANG2025110895}, a crucial task for applications such as industrial defect detection~\cite{fan2023advances}, wildlife protection~\cite{lidbetter2020search}, and medical imaging~\cite{fan2020pranet}. In static images, the main challenges arise from indistinct boundaries and background similarity, especially in complex scenes~\cite{LIU2023109514}. However, videos benefit from temporal motion cues, which enhance detection effectiveness. As a result, video camouflaged object detection (VCOD)~\cite{lamdouar2020betrayed, cheng2022implicit,hui2024implicit,hui2024endow,pang2024zoomnext} has recently garnered significant attention.

Several advanced methods have been proposed, including SAM-based VCOD techniques like TSP-SAM~\cite{hui2024endow} and MLLM-based referring expression segmentation methods such as GLaMM~\cite{rasheed2024glamm}. However, these methods have the following limitations: (1) SAM-based VCOD method~\cite{hui2024endow} struggles to separate camouflaged object edges because they freeze the SAM~\cite{kirillov2023segment} and only learn limited prompts through the self-prompt network, restricting their ability to segment edge details effectively. (2) MLLM-based referring expression segmentation methods \cite{lai2024lisa,ren2024pixellm,rasheed2024glamm} suffer from poor object separability because the LLM merges foreground and background information. Since the camouflaged object closely resembles the background, the model struggles to effectively distinguish the object from its surroundings. Figure~\ref{abstract} illustrates a conceptual diagram of these methods and their limitations in camouflaged object detection.

To address these issues, we propose a novel VCOD method based on SAM and MLLM, called \textsc{\textbf{Phantom-Insight}}, as shown in Figure~\ref{abstract}. \textit{To enhance the separability of object edge details}, we represent video sequences using both temporal and spatial clues and perform feature fusion via an LLM. Leveraging the rich pre-training knowledge of the LLM allows for better integration of information density. Subsequently, multiple cues, such as bounding boxes and masks, are generated through the dynamic foreground visual token scoring module and the prompt network to adaptively guide the SAM model, enabling it to capture subtle textures. LoRA~\cite{hu2022lora} is used to fine-tune MLLM and SAM during training.
\textit{To enhance the separability of objects and background}, we propose a decoupled foreground-background learning strategy. By generating foreground and background cues separately and conducting decoupled training, the visual token can effectively integrate foreground and background information independently. This enables SAM to more accurately segment camouflaged objects in the video. 

Experiments on the MoCA-Mask~\cite{cheng2022implicit} and CAD2016~\cite{bideau2016s} datasets demonstrate that the proposed \textsc{\textbf{Phantom-Insight}} achieves state-of-the-art performance across various evaluation metrics. On the MoCA-Mask benchmark, compared to SAM-based VCOD method TSP-SAM~\cite{hui2024endow}, our method outperforms it by 7.0\% on the $S_{\alpha}$ metric, 24.8\% on the $F_w^{\beta}$ metric, 23.4\% on the mDice metric, and 24.2\% on the mIoU metric. Our method more effectively utilizes temporal and spatial clues, along with the enhanced cues generated by the fused representation to guide and fine-tune the SAM model, enabling it to distinguish texture details more accurately. Furthermore, compared to the MLLM-based referring expression segmentation method GLaMM~\cite{rasheed2024glamm}, we surpass it by 7.4\% on the $S_{\alpha}$ metric, 22.8\% on the $F_w^{\beta}$ metric, 20.0\% on the mDice metric, and 23.3\% on the mIoU metric. With the introduction of the decoupled learning strategy, our method achieves more precise positioning of camouflaged objects. 

In summary, our main contributions can be summarized as follows:
\begin{itemize}
    \item We propose \textsc{\textbf{Phantom-Insight}}, a framework for VCOD that uses MLLM to integrate temporal-spatial information. It generates multiple cues to adaptively guide the SAM model, enabling it to accurately segment camouflaged objects.
    \item We propose a dynamic foreground visual token scoring module that effectively distinguishes between foreground and background visual tokens, enabling a more accurate generation of corresponding improvements.
    \item We propose a decoupled foreground-background learning strategy to enhance the separability of objects and background.
    \item Extensive experiments on the MoCA-Mask and CAD2016 datasets show that our method outperforms state-of-the-art methods, demonstrating its validation and generalization ability in complex camouflaged scenes.
\end{itemize}
\section{Related Work}

\subsection{Video Camouflaged Object Detection}
Video Camouflage Object Detection aims to leverage the temporal-spatial information contained in consecutive frames to identify camouflage objects that blend seamlessly with the background~\cite{yan2019semi,ji2021progressively}. 
Traditional methods \cite{lamdouar2020betrayed, cheng2022implicit} typically adopt a two-stage architecture, modeling temporal-spatial consistency through optical flow techniques or network-learned inter-frame relationships for camouflage detection. However, due to the two-stage structure, these methods are prone to error accumulation, which limits their ability to effectively capture the full extent of camouflaged objects. To address this issue, recent approaches have adopted end-to-end training architectures \cite{hui2024implicit, hui2024endow, pang2024zoomnext}. IMEX \cite{hui2024implicit} introduced a cross-scale representation fusion method that eliminates the effects of camera motion through global frame alignment and ensures consistency in camouflage object segmentation across consecutive frames using a consistency-preserving strategy. TSP-SAM \cite{hui2024endow} establishes inter-frame motion in the frequency domain, combining long-range consistency to provide reliable cues to the Segment Anything Model (SAM) for adaptive rapid learning. ZoomNeXt \cite{pang2024zoomnext} explores subtle semantic features across mixed scales, aggregating and enhancing scale-specific representations through a carefully designed perception network, while significantly suppressing background uncertainty and interference through UAL loss, thereby achieving accurate camouflage object detection. In comparison to existing VCOD methods \cite{hui2024implicit, hui2024endow, pang2024zoomnext}, our proposed \textsc{\textbf{Phantom-Insight}} integrates temporal-spatial clues within a segmentation framework combining multimodal large language model (MLLM) and Segment Anything Model (SAM), performing foreground-background decoupling learning with token-adaptive dynamic scoring via fine-tuning, thus enhancing the granularity and generalization of camouflaged object detection.

\subsection{MLLMs-based Segmentation}
With the rapid development of MLLMs~\cite{li2024llava,liu2024visual,li2023blip}, significant advancements have been made in their ability to understand both text and image content. This dual understanding enables users to query the model not only for textual information but also for specific details about visual objects, such as pinpointing the exact location of an object within an image and obtaining its pixel-level segmentation. LISA \cite{lai2024lisa} builds on this capability by introducing a segmentation token, \texttt{$<\text{SEG}>$}, which triggers the model to output an embedding that is then fed into the Segment Anything Model (SAM) to guide it in decoding the corresponding object segmentation mask. PixelLM \cite{ren2024pixellm} takes a slightly different approach, incorporating a segmentation codebook with learnable tokens that encode context and knowledge specific to the target. It also introduces a lightweight pixel decoder, which generates the target segmentation mask based on the hidden embedding of the codebook tokens and the image features. On the other hand, GLaMM \cite{rasheed2024glamm} innovates by proposing both a global image encoder and a region encoder that can process text as well as optional visual prompts (such as image-level and region-level prompts). This allows the model to interact across multiple granularity levels, enabling it to generate more precise and detailed target segmentation information. These developments collectively highlight the growing potential of MLLMs in performing complex visual tasks, offering powerful tools for a wide range of applications in computer vision and multimodal learning. However, to date, no research has focused on MLLM for video-level camouflaged object detection. To address this gap, \textsc{\textbf{Phantom-Insight}} explores temporal-spatial clues and introduces the first MLLM-based framework for video camouflaged object detection

\section{Methodology} 

\begin{figure*}[!t] 
    \centering
    \includegraphics[width=0.8\textwidth,keepaspectratio]{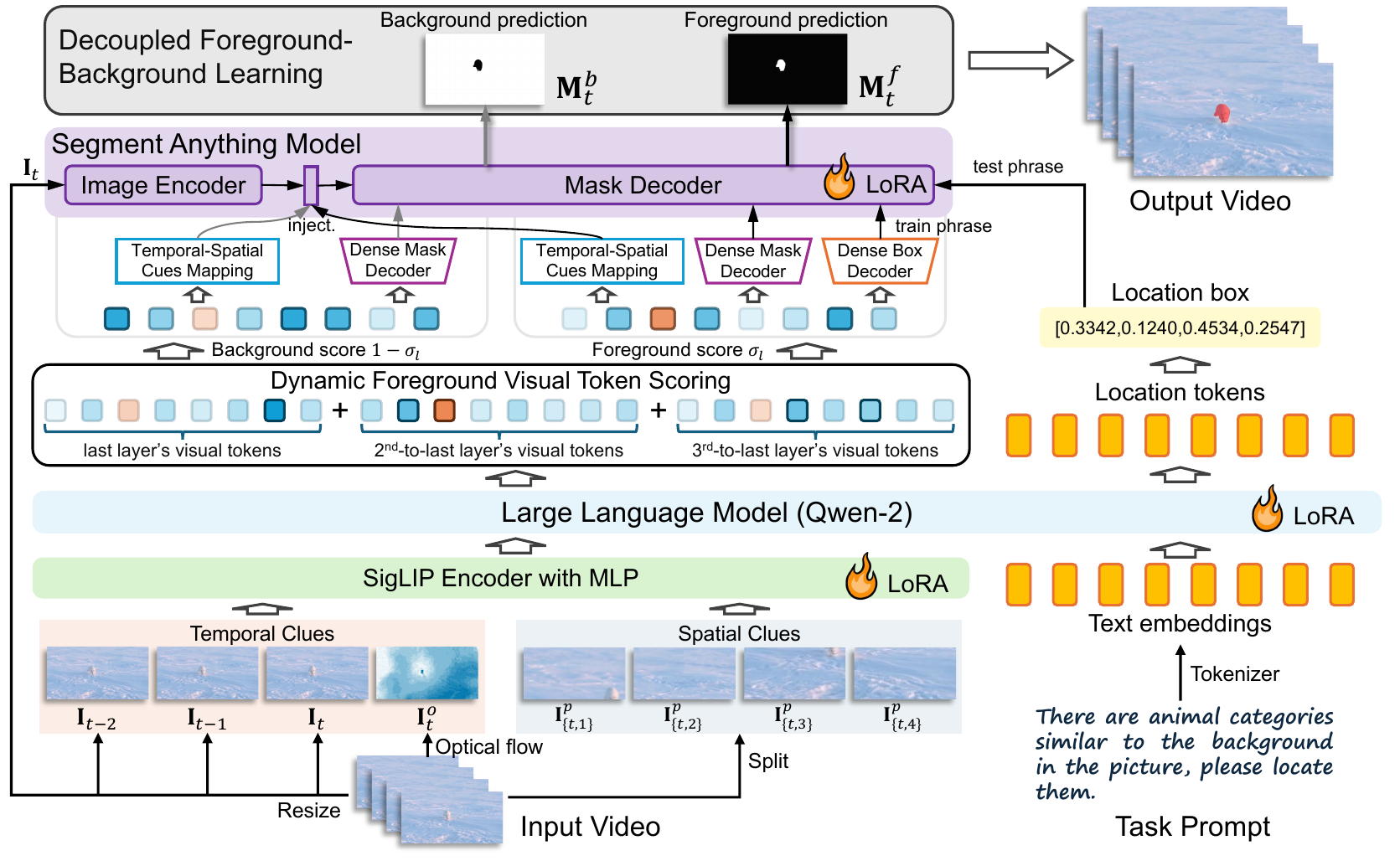} 
    \caption{The framework of the proposed \textsc{\textbf{Phantom-Insight}}. Given a video, we start predicting segmentation maps frame by frame from the third frame onward. Temporal clues, derived from consecutive frames and optical flow map, along with spatial clues generated by splitting patches using the AnyRes strategy, are combined with the task prompt and fed into the LLM to output the textual box. Subsequently, visual tokens from the last three layers undergo dynamic scoring to extract distinct tokens for foreground and background tasks, generating multiple cues for SAM to enable Decoupled Foreground-Background Learning. Note that during the inference phase, the box prompt is replaced by the textual location box predicted by the LLM, which ultimately yields the final predicted video sequence. We use LoRA to efficiently fine-tune the LLM, SigLIP, and SAM’s mask decoder.}
    \label{framework}
\end{figure*}

In this section, we present our proposed framework for video camouflaged object detection, as illustrated in Figure~\ref{framework}. 

\subsection{Temporal-spatial Enhanced Representation via MLLM}\label{sec:mllm_fusion}


We integrate the input visual representation with textual embeddings using a large language model. This approach aims to leverage the resulting visual representation for efficient camouflaged object segmentation, while the textual representation is utilized to generate location information of the camouflaged object in textual form.



\textbf{Temporal-spatial Visual Representation:} 
We incorporate multiple visual clues to enhance the visual representation, including multi-frame images and an optical flow map for capturing temporal information, as well as patch-based image splitting to emphasize details and spatial features. For a input video $\{\mathbf{I}_t\}_{t=1}^T$, where $\mathbf{I}_t \in \mathbb{R}^{H \times W \times 3}$ denote the $t$-th frame, $H$ and $W$ represent the height and width, respectively. We utilize the last three consecutive frames, $\{\mathbf{I}_{t-2}, \mathbf{I}_{t-1}, \mathbf{I}_t\}$, along with an optical flow map $\mathbf{I}^{o}_{t}$, as \textit{temporal clues}. The optical flow map is computed from the last two frames, $\mathbf{I}^{o}_{t} = \varphi(\mathbf{I}_{t-1}, \mathbf{I}_t)$, following the method described in~\cite{teed2020raft}. To enrich \textit{spatial information}, we adopt the AnyRes strategy~\cite{li2024llava}, which adaptively splits the image into $K$ patches, where the $i$-th patch is denoted as $\mathbf{I}^{p}_{\{t,i\}}$. All temporal-spatial clues are resized to $384 \times 384$ and input into a SigLip encoder $E(\cdot)$, followed by an MLP layer, to extract visual representations. Each image generates 729 visual tokens, resulting in visual representations of size $729 \times 3084$ for time-series frames $\{\mathbf{f}_{t-2},\mathbf{f}_{t-1},\mathbf{f}_{t}\}$, the optical flow map $\mathbf{f}^{o}_{t}$, and patches $\mathbf{f}^{p}_{\{t, i\}}$. The process can be summarized as follows:
\begin{equation}
    \begin{split}
    \{\mathbf{f}_{t-2},\mathbf{f}_{t-1},\mathbf{f}_{t}\} =& \{E(\mathbf{I}_{t-2}),E(\mathbf{I}_{t-1}),E(\mathbf{I}_t)\},\\
    \mathbf{f}^{o}_t =& E(\mathbf{I}^{o}_t),\;\;\;\mathbf{f}^{p}_{\{t,i\}} = E(\mathbf{I}^{p}_{\{t,i\}}).
    \end{split}
\end{equation}

During training, we apply LoRA ~\cite{hu2022lora} fine-tuning to the SigLip encoder to better adapt it to this task.

\textbf{Text Representation:} We define a sentence with special semantics to prompt the Large Language Model:

{\texttt{There are animal categories similar to the  background \\ in the image, please locate them.}}

The prompt is processed through a tokenizer and embedding layers to generate the text representation $\mathbf{f}_{\text{text}}$.

\textbf{Multimodal Representation Fusion:} We concatenate the temporal-spatial visual representations with the text representation and feed them into the Qwen-2 Large Language Model~\cite{bai2023qwen} to fuse the representations after the autoregressive process:
\begin{equation}
    \begin{split}
        & [\mathbf{f}^{\text{visual}}_{l}, \mathbf{f}^{\text{location}}] = \\
        & \mathsf{LLM}\left(\left[\underbrace{\mathbf{f}_{t-2},\mathbf{f}_{t-1},\mathbf{f}_{t},\mathbf{f}_{t}^{o}}_{\text{temporal clues}},\underbrace{\mathbf{f}_{\{t,1\}}^{p},\ldots,\mathbf{f}_{\{t,K\}}^{p}}_{\text{spatial clues}},\underbrace{\mathbf{f}_{\text{text}}}_{\text{prompt}} \right] \right),
    \end{split}
\end{equation}
where, $l$ denotes the output of the $l$-th last layer of the Large Language Model. $\mathbf{f}^{\text{visual}}_{l}$ aligns with the positions of the input temporal and spatial cue tokens, and their size corresponds to their combined size. The outputs from the last three layers are obtained and used for subsequent foreground and background prediction. $\mathbf{f}^{\text{location}}$ represents the textual descriptive features related to location information generated by the Large Language Model. These features are decoded into textual descriptions through the Large Language Model’s vocabulary classifier, facilitating the localization of object information (\textit{e.g.}, \textsc{[0.3342, 0.1240, 0.4534, 0.2547]}).

During training, we apply LoRA fine-tuning to the Large Language Model to enable it to understand the dynamic movement of objects between frames and the current spatial location.


\subsection{SAM-based Multi-cue Enhanced Segmentation}\label{sec:SAM}

After fully achieving multimodal fusion, we leverage multi-cue visual features to accurately segment camouflaged objects, with the Segment Anything Model (SAM)~\cite{kirillov2023segment} serving as the baseline.

\textbf{Dynamic Foreground Visual Token Scoring:} We obtain the visual tokens $\mathbf{f}_{l}^{\text{visual}}$ from the last three layers of the Large Language Model. Since different visual tokens capture varying degrees of foreground and background information, we adopt a dynamic approach to assess their contribution to the task. Specifically, we design a dynamic scoring network $\mathsf{MLP}_s(\cdot)$ consisting of two MLP layers, followed by a sigmoid function $\sigma(\cdot)$. The visual tokens from the last three layers are fed into the network to predict the foreground score $s_{l}$ for each layer's token, 
\begin{equation}
    s_{l} = \sigma\left(\mathsf{MLP}_s\left(\mathbf{f}_{l}^{\text{visual}}\right)\right),
\end{equation}
then, we use the scores to weigh the tokens at each layer, apply adaptive pooling $\mathsf{Pooling}(\cdot)$ to reduce the number of tokens per layer, and sum the weighted tokens to obtain the foreground token representation,
\begin{equation}
    \mathbf{f}_{\text{fg}} = \sum_{l \in \text{last 3 layers}}{\mathsf{Pooling}\left(s_{l} \cdot \mathbf{f}_{l}^{\text{visual}}\right)},
\end{equation}
which will be utilized to predict the foreground of camouflaged objects. Similarly, the background token representation, used for predicting the background, will be computed as follows:
\begin{equation}
    \mathbf{f}_{\text{bg}} = \sum_{l \in \text{last 3 layers}}{\mathsf{Pooling}\left((1 - s_{l}) \cdot \mathbf{f}_{l}^{\text{visual}}\right)}.
\end{equation}



\textbf{Multi-cue Generation:} To detect camouflaged objects, we utilize the SAM for pixel-level segmentation. To enhance SAM’s localization accuracy, we generate multiple cues by separately predicting the foreground and background. For foreground prediction, we generate a temporal-spatial cue, mask prompt, and box prompt, derived from the foreground token representations. 
To generate the temporal-spatial cue and mask prompt, adaptive pooling is first applied to obtain $\mathbf{f}^{\text{long}}_{\text{fg}}$, consisting of 256 tokens. A two-layer MLP network $\mathsf{MLP}_{\text{cue}}(\cdot)$ and a mask decoder $\mathsf{Decoder}_{M}(\cdot)$ are then employed to produce the temporal-spatial cue feature $\mathbf{f}^{c}_{\text{fg}}$ and the mask prompt $\mathbf{p}^{\text{mask}}_{\text{fg}}$, respectively:
\begin{equation}
    \mathbf{f}_{\text{fg}}^{c} = \mathsf{MLP}_{\text{cue}}\left(\mathbf{f}^{\text{long}}_{\text{fg}}\right),~~~~\mathbf{p}^{\text{mask}}_{\text{fg}} = \mathsf{Decoder}_{M}\left(\mathbf{f}^{\text{long}}_{\text{fg}}\right),
\end{equation}
then, to generate the box prompt, adaptive pooling is applied to obtain $\mathbf{f}^{\text{short}}_{\text{fg}}$, consisting of 4 tokens. Using a box decoder $\mathsf{Decoder}_{B}(\cdot)$, composed of two MLP layers with ReLU activation, the box prompt is obtained as:
\begin{equation}
    \mathbf{p}^{\text{box}}_{\text{fg}} = \mathsf{Decoder}_{B}\left(\mathbf{f}^{\text{short}}_{\text{fg}}\right).
\end{equation}
For background prediction, we generate a temporal-spatial cue and a mask prompt, derived from the background token representations using the shared network employed for foreground cue generation:
\begin{equation}
    \mathbf{f}_{\text{bg}}^{c} = \mathsf{MLP}_{\text{cue}}\left(\mathbf{f}_{\text{bg}}\right),~~~~\mathbf{p}^{\text{mask}}_{\text{bg}} = \mathsf{Decoder}_{M}\left(\mathbf{f}_{\text{bg}}\right).
\end{equation}
Next, we utilize the generated multi-cue and the SAM for segmentation.


\textbf{SAM-based Segmentation:} 
We segment the frame image at time $t$ by inputting $\mathbf{I}_{t}$ into SAM’s image encoder $\mathsf{SAM}_{E}(\cdot)$. For foreground prediction, the temporal-spatial cue $\mathbf{f}_{\text{fg}}^{c}$ is integrated into the image features encoded by SAM to enhance its temporal-spatial perception. Simultaneously, the generated mask prompt $\mathbf{p}^{\text{mask}}_{\text{fg}}$ and box prompt $\mathbf{p}^{\text{box}}_{\text{fg}}$ are jointly input into SAM’s mask decoder $\mathsf{SAM}_{D}(\cdot)$ to guide the model, ultimately producing the pixel-level foreground prediction $\mathbf{M}_{t}^{f}$:
\begin{equation}
    \mathbf{M}_{t}^{f} = \mathsf{SAM}_{D} \left( \mathsf{SAM}_{E}\left(\mathbf{I}_{t}\right) + \mathbf{f}_{\text{fg}}^{c}, \mathbf{p}^{\text{mask}}_{\text{fg}}, \mathbf{p}^{\text{box}}_{\text{fg}} \right ),
\end{equation}
note that during the inference stage, the box information corresponding to the text representation $\mathbf{f}^{\text{location}}$ is directly used to replace $\mathbf{p}^{\text{box}}_{\text{fg}}$.
For background prediction, the temporal-spatial cue $\mathbf{f}_{\text{bg}}^{c}$ is injected into the image features encoded by SAM. The generated mask prompt $\mathbf{p}^{\text{mask}}_{\text{bg}}$ is then used to guide the model in segmenting the pixel-level background:
\begin{equation}
    \mathbf{M}_{t}^{b} = \mathsf{SAM}_{D} \left( \mathsf{SAM}_{E}\left(\mathbf{I}_{t}\right) + \mathbf{f}_{\text{bg}}^{c}, \mathbf{p}^{\text{mask}}_{\text{bg}} \right ).
\end{equation}

During the training process, we share the same SAM model weights for both tasks and perform LoRA fine-tuning on SAM’s mask decoder. This allows the model to quickly adapt to the camouflaged object segmentation task and accelerates convergence.


\subsection{Decoupled Foreground-Background Learning}\label{sec:decoupled_learning}

Since the foreground and background information of the camouflaged object may become coupled during Large Language Model fusion, we adopt a learning approach that predicts the foreground and background separately. Additionally, we propose a decoupled foreground-background learning strategy.

For a given ground truth mask $ \mathbf{m}_{\text{gt}} $ and a predicted mask $ \mathbf{m}_{\text{pred}} $, as well as a ground truth box $\boldsymbol{b}_{\text{gt}} $ and a predicted box $\boldsymbol{b}_{\text{pred}} $, the segmentation loss, which includes Binary Cross-Entropy (BCE) loss and Dice loss, and the box loss, consisting of L1 loss and Generalized Intersection over Union (GIoU) loss, are defined as follows:
\begin{align}
\mathcal{L}_{\text{seg}} =&~\mathsf{BCE}\left(\mathbf{m}_{\text{gt}},\mathbf{m}_{\text{pred}}\right) + \mathsf{Dice}\left(\mathbf{m}_{\text{gt}}, \mathbf{m}_{\text{pred}}\right),\\
\mathcal{L}_{\text{box}} =&~\mathsf{L1}\left(\boldsymbol{b}_{\text{gt}}, \boldsymbol{b}_{\text{pred}}\right) + \mathsf{GIoU}\left(\boldsymbol{b}_{\text{gt}}, \boldsymbol{b}_{\text{pred}}\right).
\end{align}

For the foreground task, we utilize masks and bounding boxes of specific objects as prompts for SAM. As a result, we incorporate a segmentation loss to guide the mask decoder in generating the foreground prompt mask $\mathbf{p}^{\text{mask}}_{\text{fg}}$. Additionally, we introduce bounding box losses to supervise the box decoder in generating the foreground box prompt $\mathbf{p}^{\text{box}}_{\text{fg}}$.
In contrast, for the background task, only masks are used as prompts for SAM. Therefore, we introduce only a segmentation loss to guide the shared-weight mask decoder in generating the background mask prompt  $\mathbf{p}^{\text{mask}}_{\text{bg}}$. The overall prompt loss is defined as follows:
\begin{equation}
    \begin{aligned}
        \mathcal{L}_{\text{prompt}} =&\mathcal{L}_{\text{box}}\left(\mathbf{p}^{\text{box,gt}}_{\text{fg}},\mathbf{p}^{\text{box}}_{\text{fg}}\right) +  \mathcal{L}_{\text{seg}}\left(\mathbf{M}_{t}^{f,gt}, \mathbf{p}^{\text{mask}}_{\text{fg}}\right) \\
        &+\mathcal{L}_{\text{seg}}\left(\mathbf{M}_{t}^{b,gt}, \mathbf{p}^{\text{mask}}_{\text{bg}}\right).
    \end{aligned}
\end{equation}

We employ a cross-entropy loss to train Large Language Model to generate predicted textual bounding boxes $b_{\text{pred}}^{\text{text}} $, and the loss function is defined as follows:
\begin{equation}
\mathcal{L}_{\text{text}} = \mathsf{CE}\left(b_{\text{gt}}^{\text{text}}, b_{\text{pred}}^{\text{text}}\right),
\end{equation}
where $b_{\text{gt}}^{\text{text}}$ is the ground truth box information.

For the foreground prediction, we supervise the final generated foreground mask $\mathbf{M}_{t}^{f}$ using a segmentation loss. Similarly, for the background prediction, we supervise the final generated background mask $\mathbf{M}_{t}^{b}$ with a segmentation loss, as shown below:
\begin{equation}
\mathcal{L}_{\text{mask}} = \mathcal{L}_{\text{seg}}\left(\mathbf{M}_{t}^{f,gt},\mathbf{M}_{t}^{f}\right)+\mathcal{L}_{\text{seg}}\left(\mathbf{M}_{t}^{b,gt},\mathbf{M}_{t}^{b}\right).
\end{equation}

In summary, the overall optimization objective is defined as follows:
\begin{equation}
\mathcal{L} = \mathcal{L}_{\text{prompt}} + \mathcal{L}_{\text{text}} + \mathcal{L}_{\text{mask}},
\end{equation}
where, $\mathcal{L}_{\text{prompt}}$ and $\mathcal{L}_{\text{text}}$ will optimize the parameters of the Large Language Model, SigLIP, and the related mapping layers, while $\mathcal{L}_{\text{mask}}$ will further optimize the parameters of the SAM decoder.

\subsection{Inference}\label{sec:inference}

Our inference process consists of four main steps:
(1) Input the video, skipping the first two frames. Starting from the third frame, we predict the camouflaged object frame by frame.
(2) Use the current frame, along with the previous two frames and the optical flow map, as temporal clues, the split patches of the current frame are used as spatial clues. Then fed them into the SigLip encoder with an MLP to obtain visual representations.
(3) Use the Large Language Model to fuse the visual representations with the text representation, obtaining foreground multiple cues and textual box, and then use SAM to predict the foreground segmentation map.
(4) The predicted foreground segmentation maps for each frame are stitched together to produce the final result of video camouflage target detection.

\section{Experiments}


\subsection{Experimental Setup}\label{experimental_setup}

\textbf{Datasets.} 
We use two publicly available video camouflaged object detection datasets: MoCA-Mask~\cite{cheng2022implicit} and CAD2016~\cite{bideau2016s}. MoCA-Mask is the largest dataset for camouflaged animal detection, containing 87 videos (22,939 frames) with bounding box and pixel-level annotations every five frames. It is split into 71 videos (19,313 frames) for training and 16 videos (3,626 frames) for testing. CAD2016 includes 9 YouTube videos with 181 pixel-level masks, also annotated every five frames. For consistent evaluation, we train on the 71 MoCA-Mask videos and test on the remaining MoCA-Mask and all CAD2016 sequences.

\textbf{Evaluation Metrics.} 
We adopt six standard metrics to evaluate the VCOD task: (1) S-measure $(S_{\alpha})$~\cite{fan2017structure} for structural similarity, combining region and object-aware cues; (2) Weighted F-measure $(F_w^{\beta})$~\cite{margolin2014evaluate}, balancing precision and recall via a tunable $\beta$; (3) Enhanced-alignment measure $(E_{\phi})$~\cite{fan2021cognitive}, assessing both pixel-wise alignment and global statistics; (4) Mean Absolute Error (MAE, $\mathcal{M}$), computing average pixel-wise differences; (5) Mean Dice (mDice), evaluating mask overlap using the Dice coefficient; and (6) Mean IoU (mIoU), measuring intersection-over-union. Together, these metrics offer a comprehensive evaluation of structural, pixel-level, and boundary performance.

\begin{table*}[h]
    \centering
    \caption{Video camouflaged object detection evaluation results on the MoCA-Mask~\cite{cheng2022implicit} and CAD2016~\cite{bideau2016s} datasets. \textbf{Bold} indicates the highest performance, and \underline{underlined} represents the second highest performance.}
    \resizebox{\textwidth}{!}{
    \begin{tabular}{lc|cccccc|cccccc}
    \toprule
    \multirow{2}{*}{Methods}          & \multirow{2}{*}{Input Modality}     & \multicolumn{6}{c}{MoCA-Mask~\cite{cheng2022implicit}}                                                                                    & \multicolumn{6}{c}{CAD2016~\cite{bideau2016s}}                                                                         \\ \cmidrule(l){3-14} 
                                     &                            & $S_{\alpha}\uparrow$             & $F_w^{\beta}\uparrow$             & $E_{\phi}\uparrow$             & $\mathcal{M}\downarrow$             & mDice $\uparrow$         & mIoU $\uparrow$  & $S_{\alpha}\uparrow$             & $F_w^{\beta}\uparrow$             & $E_{\phi}\uparrow$             & $\mathcal{M}\downarrow$             & mDice $\uparrow$         & mIoU $\uparrow$          \\ \midrule
    \multicolumn{14}{c}{Single-image camouflaged object detection}  \\ \midrule
    SINet~\cite{fan2020camouflaged} & \multirow{6}{*}{Image} & 0.574          & 0.185          & 0.655          & 0.030          & 0.221          & 0.156  & 0.636          & 0.346          & 0.775          & 0.041          & 0.381          & 0.283          \\
    SINet-v2~\cite{fan2021concealed}   &  & 0.571          & 0.175          & 0.608          & 0.035          & 0.211          & 0.153  & 0.653          & 0.382          & 0.762          & 0.039          & 0.413          & 0.318          \\
    ZoomNet~\cite{pang2022zoom} &  & 0.582          & 0.211          & 0.536          & 0.033          & 0.224          & 0.167  & 0.633          & 0.349          & 0.601          & 0.033          & 0.349          & 0.273          \\
    FEDERNet~\cite{he2023camouflaged} &  & 0.555          & 0.158          & 0.542          & 0.049          & 0.192          & 0.132  & 0.573          & 0.272          & 0.609          & 0.051          & 0.260          & 0.199          \\
    FSPNet~\cite{huang2023feature} &  & 0.594          & 0.182          & 0.608          & 0.044          & 0.238          & 0.167  & 0.681          & 0.401          & 0.716          & 0.044          & 0.446          & 0.332          \\
    PUENet~\cite{zhang2023predictive} &  & 0.594          & 0.204          & 0.619          & 0.037          & 0.302          & 0.212  & 0.691          & 0.485          & 0.795          & 0.034          & 0.514          & 0.396          \\ \midrule
    \multicolumn{14}{c}{MLLM-based referring expression segmentation}  \\ \midrule
    LISA$^{*}$~\cite{lai2024lisa}    & \multirow{6}{*}{Image} & 0.592         & 0.263         & 0.741         & 0.016         & 0.254         & \multicolumn{1}{c|}{0.219} & 0.661         & 0.437         & 0.686         & 0.032         & 0.374         & 0.305         \\
    LISA$^{\dagger}$~\cite{lai2024lisa}   &  & 0.546         & 0.178         & 0.685         & 0.033         & 0.177         & 0.148 & 0.651         & 0.409         & 0.774         & 0.041         & 0.368         & 0.293         \\
    PixelLM$^{*}$~\cite{ren2024pixellm} &  & 0.587         & 0.279         & 0.669         & 0.012         & 0.259         & 0.196 & 0.601         & 0.377         & 0.654         & 0.127         & 0.397         & 0.306         \\
    PixelLM$^{\dagger}$~\cite{ren2024pixellm} &  & 0.591         & 0.304         & 0.647         & 0.011         & 0.269         & 0.199 & 0.662         & 0.461         & 0.680         & 0.031         & 0.395         & 0.302         \\
    GLaMM$^{*}$~\cite{rasheed2024glamm} &  & 0.511         & 0.150         & 0.544         & 0.087         & 0.163         & 0.127 & 0.657         & 0.438         & 0.744         & 0.045         & 0.403         & 0.337         \\
    GLaMM$^{\dagger}$~\cite{rasheed2024glamm} &  & 0.686         & 0.451         & 0.802         & 0.015         & 0.471         & 0.391 & 0.718         & 0.557         & \underline{0.873}  & 0.030         & 0.534         & 0.438         \\ \midrule
    \multicolumn{14}{c}{Video-based camouflaged object detection}                              \\ \midrule
    RCRNet~\cite{yan2019semi} & \multirow{8}{*}{Video} & 0.597          & 0.174          & 0.583          & 0.025          & 0.194          & \multicolumn{1}{c|}{0.137}  & 0.627          & 0.287          & 0.666          & 0.048          & 0.309          & 0.229          \\
    PNS-Net~\cite{ji2021progressively} &  & 0.576          & 0.134          & 0.562          & 0.038          & 0.189          & 0.133  & 0.655          & 0.325          & 0.673          & 0.048          & 0.384          & 0.290          \\
    MG~\cite{yang2021self} &  & 0.547          & 0.165          & 0.537          & 0.095          & 0.197          & 0.141  & 0.594          & 0.336          & 0.691          & 0.059          & 0.368          & 0.268          \\
    SLT-Net~\cite{cheng2022implicit} &  & 0.656          & 0.357          & 0.785          & 0.021          & 0.387          & 0.310  & 0.696          & 0.481          & 0.845          & 0.030          & 0.493          & 0.401          \\
    IMEX~\cite{hui2024implicit} &  & 0.661          & 0.371          & 0.778          & 0.020          & 0.409          & \multicolumn{1}{c|}{0.319}  & 0.695          & 0.490          & 0.846          & 0.030          & 0.501          & 0.412          \\
    SAM-PM~\cite{meeran2024sam} &  & 0.695          & 0.464          & 0.732          & 0.011          & \underline{0.497}          & \multicolumn{1}{c|}{0.416}  & 0.729          & 0.602          & 0.746          & \textbf{0.018} & 0.594          & 0.493          \\
    TSP-SAM~\cite{hui2024endow} &  & 0.689          & 0.444          & \underline{0.808}          & \textbf{0.008} & 0.458          & 0.388  & 0.751          & \underline{0.628}          & 0.851          & 0.021          & \underline{0.603}          & 0.496          \\
    ZoomNeXt~\cite{pang2024zoomnext}  &  & \underline{0.734} & \underline{0.476}          & 0.736          & \underline{0.010}          & 0.490          & \underline{0.422}  & \underline{0.757}          & 0.593          & 0.865          & 0.020          & 0.599          & \underline{0.510}  \\ \midrule
    \rowcolor{gray!20}
    \textsc{\textbf{Phantom-Insight}} (Ours)         & Video                      & \textbf{0.737}         & \textbf{0.554} & \textbf{0.836} & 0.011 & \textbf{0.565} & \textbf{0.482}              & \textbf{0.783} & \textbf{0.640} & \textbf{0.888} & \underline{0.019} & \textbf{0.636} & \textbf{0.541} \\ \bottomrule
    \end{tabular}}
    \label{benchmark_comparison}
    \begin{tablenotes} 
        \small
        \item $*$ denotes the zero-shot capability of MLLM and $\dagger$ denotes the fine-tuning of MLLM on the MoCA-Mask dataset.
    \end{tablenotes} 
\end{table*}

\textbf{Baselines.} 
To evaluate the effectiveness of the proposed method, we compare it with several state-of-the-art approaches, grouped into three distinct categories: (1) \textit{\underline{Single-image camouflaged object} \\ \underline{detection methods}} focus on static images to help the model identify subtle differences between camouflaged objects and their surroundings from diverse perspectives. Representative methods include SINet \cite{fan2020camouflaged}, SINet-v2 \cite{fan2021concealed}, ZoomNet \cite{pang2022zoom}, BGNet \cite{yan2019semi}, FEDERNet \cite{he2023camouflaged}, FSPNet \cite{huang2023feature}, and PUENet \cite{zhang2023predictive}. (2) \textit{\underline{MLLM-based referring expression}\\ \underline{ segmentation methods}} leverage MLLMs to perform segmentation. Given a query like \texttt{"Can you segment the camouflaged animal in this image?"}, the LLM replies \texttt{"Sure, the segmentation is $<\text{SEG}>$"}, using the embedding of \texttt{$<\text{SEG}>$} to guide the mask decoder. These models benefit from extensive pretraining and fine-tuning on visual grounding tasks. We adopt LISA \cite{lai2024lisa}, PixelLM \cite{ren2024pixellm}, and GLaMM \cite{rasheed2024glamm} as baselines.(3) \textit{\underline{Video-based camouflaged object }\\ \underline{detection methods}} exploit temporal-spatial clues across consecutive frames to locate and track camouflaged objects for precise segmentation. Notable methods include RCRNet \cite{yan2019semi}, PNS-Net \cite{ji2021progressively}, MG \cite{yang2021self}, SLT-Net \cite{cheng2022implicit}, IMEX \cite{hui2024implicit}, SAM-PM \cite{meeran2024sam}, TSP-SAM \cite{hui2024endow}, and ZoomNeXt \cite{pang2024zoomnext}.

\textbf{Implementation Details.} During the training phase, the Qwen-2, SigLip encoder, and mask decoder of SAM are fine-tuned using LoRA with $r=128$. The remaining smaller parameter components are fully updated during the training process. The images are resized to $384 \times 384$ for both training and testing. We utilize DeepSpeed for acceleration as the training framework, with the process conducted for 2 epochs on 4 A100 GPUs, taking 8 hours. We use AdamW as the optimizer and apply the warmup and decay strategy to adjust the learning rate. The final learning rate is set to 0.0002. The batch size for a single GPU is 1 and the gradient accumulation step is set to 4.

\subsection{Evaluation on MoCA-Mask}\label{exp:moca}



We first validate our proposed \textsc{\textbf{Phantom-Insight}} on the MoCA-Mask dataset~\cite{cheng2022implicit}. As shown in Table \ref{benchmark_comparison}, our method achieves state-of-the-art performance on most evaluation metrics. Compared to the SOTA video-based method ZoomNeXt~\cite{pang2024zoomnext}, our method outperforms it by 16.3\% on the $F_w^{\beta}$ metric, 15.3\% on the mDice metric, and 10.6\% on the mIoU metric. Although ZoomNeXt also enhances temporal-spatial cues at the feature layer, it requires careful design of each network module and various loss functions. In contrast, our method extracts temporal and spatial clues at the input level and fuses them through LLM to enhance temporal-spatial cues, offering a more convenient and efficient approach. Compared to TSP-SAM~\cite{hui2024endow}, our method outperforms it by 7.0\% on the $S_{\alpha}$ metric, 24.8\% on the $F_w^{\beta}$ metric, 23.4\% on the mDice metric, and 24.2\% on the mIoU metric. Although both temporal and spatial clues are utilized, our approach holds an advantage due to its more effective use of temporal and spatial clues using LLM, as well as the enhanced cues generated by the fused representation to guide and fine-tune the SAM model.

\begin{figure}[!t] 
    \centering
    \includegraphics[width=0.48\textwidth]{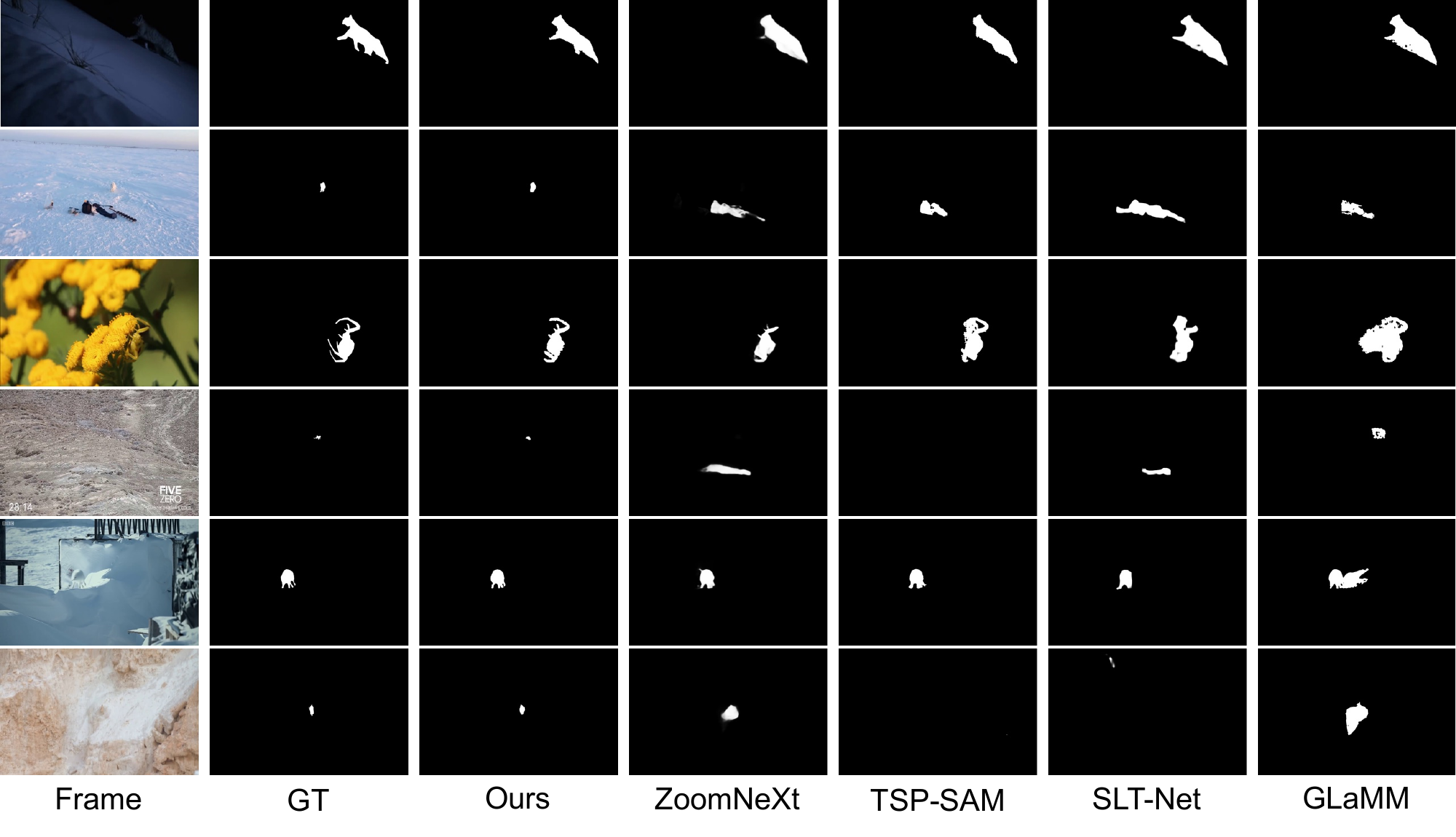} 
    \caption{Visualization of our method and baseline methods on MoCA-Mask dataset. From left to right: frame (1st column), ground truth (2nd column), prediction of our method (3rd column), and predictions of compared methods (4th-7th columns).}
    \label{vis-moca}
\end{figure}

\begin{figure}[!t] 
    \centering
    \includegraphics[width=0.48\textwidth]{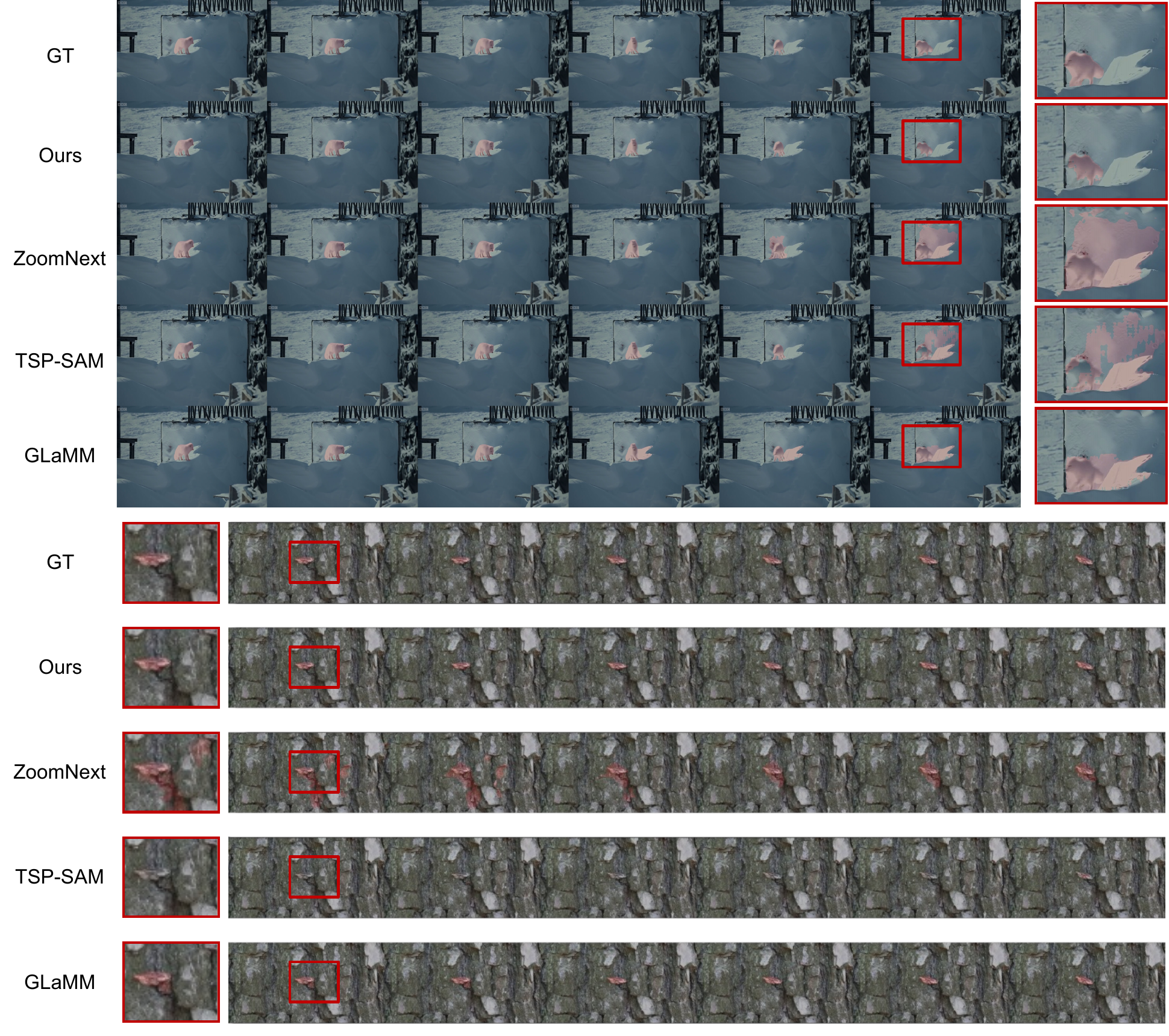} 
    \caption{Continuous multi-frame visualization of our method and baseline methods on the MoCA dataset. }
    \label{video}
\end{figure}

Compared to single-image camouflaged object detection methods, our approach significantly outperforms them across all evaluation metrics, demonstrating that the use of LLM and SAM, with their advanced pre-trained knowledge, greatly enhances the ability to detect camouflaged objects. 
Compared to MLLM-based referring expression segmentation methods, we report their zero-shot segmentation capabilities for camouflaged objects, as well as their performance after fine-tuning. Our method also achieves state-of-the-art performance regardless of whether they are fine-tuned or not. Compared to the fine-tuned GLaMM~\cite{rasheed2024glamm}, we surpass it by 7.4\% on the $S_{\alpha}$ metric, 22.8\% on the $F_w^{\beta}$ metric, 20.0\% on the mDice metric, and 23.3\% on the mIoU metric. Although GLaMM also leverages SAM and is fine-tuned on a larger number of referring expression segmentation (RES) datasets, with access to more knowledge, our method effectively fuses temporal and spatial clues and learns foreground-background decoupling, which more effectively guides SAM to achieve better performance in camouflaged object segmentation.

\begin{figure}[!t] 
    \centering
    \includegraphics[width=0.48\textwidth]{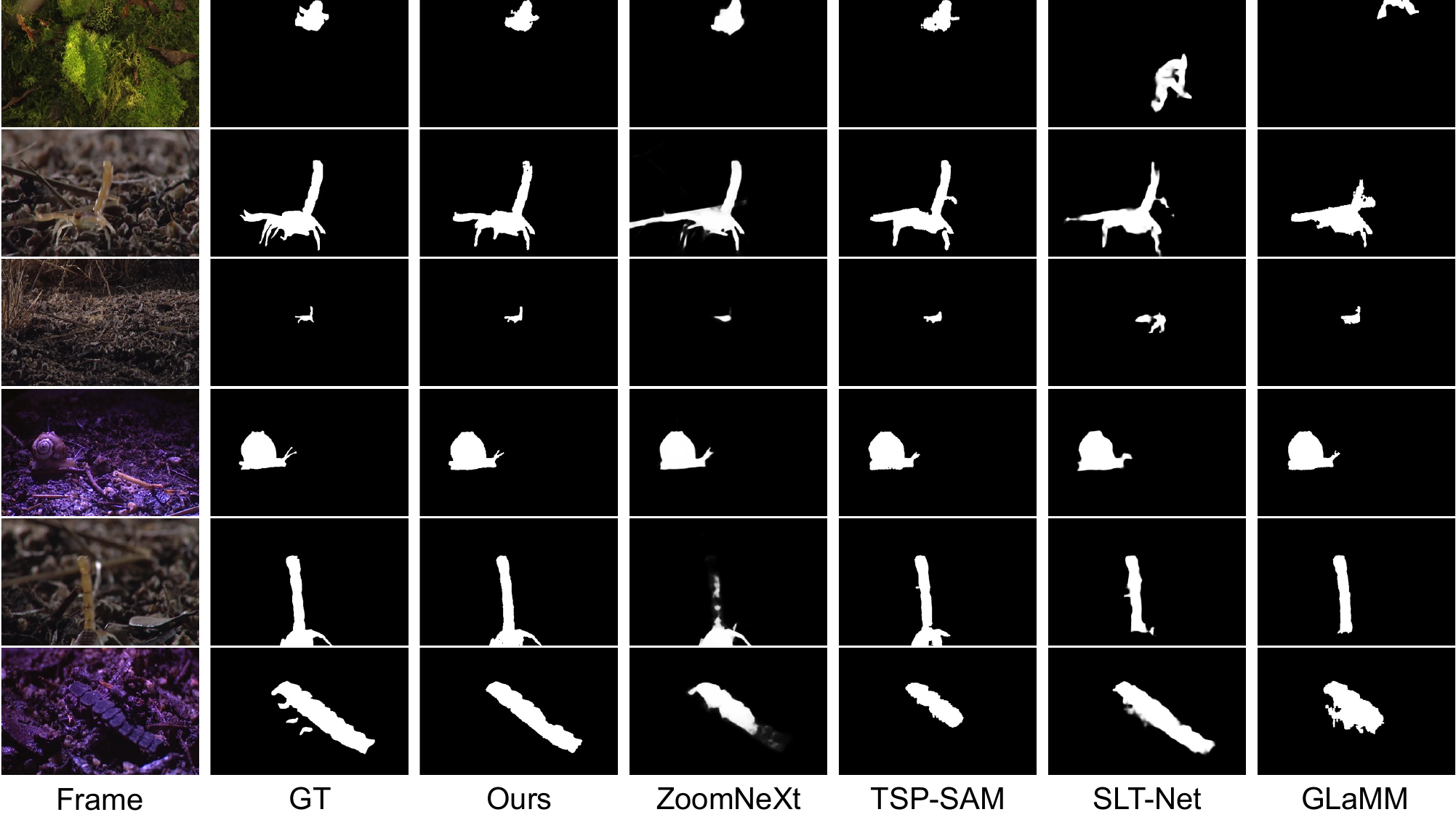} 
    \caption{Visualization of our method and baseline methods on the CAD2016 dataset. From left to right: the frame (1st column), ground truth (2nd column), prediction of our method (3rd column), and predictions of the compared methods (4th to 7th columns).}
    \label{vis-cad}
\end{figure}

To provide a more intuitive evaluation of the performance of our proposed method compared to others, we visualize a range of complex and challenging scenarios from the MoCA-Mask dataset, as shown in Figure~\ref{vis-moca}. These scenarios encompass challenging conditions such as low-light environments (Row 1), interference from multiple objects (Rows 2 and 3), ambiguous object appearances (Row 5), and small object detection difficulties (Rows 4 and 6). In each of these cases, our method effectively detects and accurately segments camouflaged objects, even under the most demanding circumstances. This performance underscores the method’s exceptional ability to extract and recognize camouflaged objects by leveraging temporal-spatial clues for object detection. Additionally, it benefits from foreground-background decoupling learning, which enables the model to successfully identify the camouflaged object despite interference from multiple salient targets in the background, remaining unaffected by these distractions.

Furthermore, as illustrated in Figure \ref{video}, we present the visualization results of continuous multi-frame object movement. Clearly, even when the object is continuously moving, our method maintains clear segmentation boundaries, whereas ZoomNeXt and TSP-SAM may predict excessive background or fail to identify the camouflaged object altogether. This further confirms the superiority of our framework in extracting and utilizing temporal-spatial representation.

\begin{table*}[!t] 
    \centering
    \caption{Ablation study of different cue generation components for SAM using the MoCA-Mask dataset.}
    \resizebox{\textwidth}{!}{ 
    \begin{tabular}{ccc|c|cccccc}
    \toprule
    \multicolumn{3}{c|}{Components for Generate Cues} & \multirow{3}{*}{Background Learning} & \multicolumn{6}{c}{Evaluation Metrics} \\  
    \begin{tabular}[c]{c}Dense Box Decoder\\ ($\mathsf{Decoder}_{B}$) \end{tabular}
 & 
\begin{tabular}[c]{c}Dense Mask Decoder\\ ($\mathsf{Decoder}_{M}$)\end{tabular}
 & 
\begin{tabular}[c]{c}Temporal-Spatial Cues Mapping\\ ($\mathsf{MLP}_{\text{cue}}$)\end{tabular}
 &  & $S_{\alpha}\uparrow$ & $F_w^{\beta}\uparrow$ & $E_{\phi}\uparrow$ & $\mathcal{M}\downarrow$ & mDice$\uparrow$ & mIoU$\uparrow$ \\ \midrule
    \Checkmark & \XSolidBrush & \XSolidBrush & \XSolidBrush & 0.680 & 0.443 & 0.768 & 0.014 & 0.453 & 0.369 \\
    \Checkmark & \Checkmark & \XSolidBrush & \XSolidBrush & 0.687 & 0.468 & 0.771 & 0.014 & 0.464 & 0.375 \\
    \Checkmark & \Checkmark & \Checkmark & \XSolidBrush & 0.706 & 0.493 & 0.805 & 0.015 & 0.499 & 0.421 \\ 
    \rowcolor{gray!20}
    \Checkmark & \Checkmark & \Checkmark & \Checkmark & \textbf{0.737} & \textbf{0.554} & \textbf{0.836} & \textbf{0.011} & \textbf{0.565} & \textbf{0.482} \\ \bottomrule
    \end{tabular}}
    \label{key components}
\end{table*}

\subsection{Evaluation on CAD2016}\label{exp:cad}

Next, we evaluate the performance of each model in terms of generalization to the CAD2016~\cite{bideau2016s} dataset after training on the MoCA-Mask dataset. As shown in Table~\ref{benchmark_comparison}, our method achieved state-of-the-art results, outperforming the ZoomNeXt~\cite{pang2024zoomnext} by 3.4\% on the $S_{\alpha}$ metric, 2.6\% on the $E_{\phi}$ metric, and 5.7\% on the mIoU metric. Additionally, our method surpassed the TSP-SAM~\cite{hui2024endow} method by 1.9\% on the $F_w^{\beta}$ metric and 5.5\% on the mIoU metric. The introduction of LLM and SAM plays a crucial role in enhancing model performance and generalization. By integrating LLM, which effectively combines temporal-spatial clues, the representations are significantly strengthened. This fusion leads to the generation of more effective and robust cues, thereby enhancing the generalization ability of SAM, even in previously unseen camouflaged object scenarios.

Figure \ref{vis-cad} presents visualizations on the CAD2016 dataset, where we select a diverse set of challenging scenarios, including objects with intricate details (Rows 1, 2, and 4), low-light scenes (Rows 4 and 6), and small target detection (Row 3). Notably, our method exhibits strong robustness in generalizing to these previously unseen scenes and objects. It consistently captures the clear contours and fine details of camouflaged objects, demonstrating its capability to effectively tackle a wide range of complex detection tasks.

\begin{table}[!t]
    \centering
    \caption{Ablation study of different visual input clues on MoCA-Mask dataset.}
    \resizebox{\linewidth}{!}{
    \begin{tabular}{c|cccccc}
    \toprule
    Visual Input Clues     &$S_{\alpha}\uparrow$             & $F_w^{\beta}\uparrow$              & $E_{\phi}\uparrow$            & $\mathcal{M}\downarrow$          & mDice$\uparrow$          & mIoU$\uparrow$          \\ \midrule
    $\mathbf{f}_{t}$         & 0.689          & 0.453          & 0.794          & 0.018         & 0.475          & 0.396          \\
    $\mathbf{f}_{t},\underbrace{\mathbf{f}_{\{t,1\}}^{p},\ldots,\mathbf{f}_{\{t,K\}}^{p}}_{\text{spatial clues}}$   & 0.707          & 0.497          & 0.798          & 0.015        & 0.514         & 0.432          \\
    \rowcolor{gray!20}
    $\underbrace{\mathbf{f}_{t-2},\mathbf{f}_{t-1},\mathbf{f}_{t},\mathbf{f}_{t}^{o}}_{\text{temporal clues}},\underbrace{\mathbf{f}_{\{t,1\}}^{p},\ldots,\mathbf{f}_{\{t,K\}}^{p}}_{\text{spatial clues}}$  & \textbf{0.737} & \textbf{0.554} & \textbf{0.836} & \textbf{0.011} & \textbf{0.565} & \textbf{0.482}          \\
    \bottomrule
    \end{tabular}}
    \label{albtion_input}
\end{table}

\subsection{Ablation Studies}\label{exp:ablation_study}

In this section, we conduct an ablation study to evaluate the impact of different visual input clues and fusion strategies, cue generation components for SAM, emphasizing their contributions. For hyperparameter ablations, please see the \textit{supplementary material}.

\subsubsection{Ablation of Visual Input Clues}

We studied the impact of different visual input clues, as shown in Table~\ref{albtion_input}. Our findings indicate that, based on the original image representation $\mathbf{f}_{t}$, introducing spatial clues improves the $ F_w^{\beta} $ metric by 9.7\%, and enhances the mDice and mIOU metrics by 8.2\% and 9.1\%, respectively. This suggests that spatial clues compensate for the spatial information lost when the original image is resized, enabling the model to more accurately perceive the location of the camouflaged object. Furthermore, by incorporating temporal clues, we observed that the model can detect the dynamic changes of the camouflaged object in real-time, which allows it to more effectively segment the specific details of the object. As a result, the $ S_{\alpha} $ and $ E_{\phi} $ metrics improved by 4.2\% and 11.5\%, respectively, while the mDice and mIoU metrics increased by 9.9\% and 11.6\%. These results confirm that our large language model leverages its contextual learning abilities to effectively extract temporal-spatial information.

In addition, we also explored the feature fusion strategy, as shown in Table~\ref{albtion_fusion}. Our results indicate that the fusion strategy based on sequence concatenation outperforms channel concatenation. It is evident that all metrics have significantly improved, demonstrating that, while the channel concatenation strategy reduces the number of tokens, it leads to the loss of important temporal-spatial information during the learning process of the large language model.

\begin{table}[!t]
    \centering
    \caption{Ablation study of feature fusion strategy on MoCA-Mask dataset.}
    \resizebox{\linewidth}{!}{
    \begin{tabular}{c|cccccc}
    \toprule
    Feature Fusion Strategy  &$S_{\alpha}\uparrow$             & $F_w^{\beta}\uparrow$              & $E_{\phi}\uparrow$            & $\mathcal{M}\downarrow$          & mDice$\uparrow$          & mIoU$\uparrow$          \\ \midrule
    Concatenation along the channel dimension  & 0.654          & 0.407          & 0.734          & 0.018          & 0.395          & 0.323          \\
    \rowcolor{gray!20}
    Concatenation along the sequence dimension  & \textbf{0.737} & \textbf{0.554} & \textbf{0.836} & \textbf{0.011} & \textbf{0.565} & \textbf{0.482} \\ \bottomrule
    \end{tabular}}
    \label{albtion_fusion}
\end{table}

\subsubsection{Ablation of SAM Cues}

We examine the impact of different cue generation components used to prompt SAM and investigate whether background prediction is simultaneously performed in camouflaged object detection, as shown in Table~\ref{key components}. We find that, based on the dense box decoder, introducing mask cues generated by the dense mask decoder improves the $F^{\beta}_{w}$ metric by 5.6\% and the mDice metric by 2.4\%, demonstrating that mask cues contribute to more precise region segmentation. After incorporating temporal-spatial cues, the model is better able to leverage contextual information, enabling it to capture local details more effectively, which improves the $E_{\phi}$ metric by 4.4\%. Additionally, it demonstrates greater accuracy in detecting camouflaged objects, with the mDice and mIoU metrics increasing by 7.5\% and 12.3\%, respectively. Finally, we explore simultaneous background prediction to disentangle the model’s mixing of foreground and background information. We find that this strategy further enhances the performance of video camouflaged object detection. It not only more effectively distinguishes the subtle differences between foreground and background, improving the $S_{\alpha}$ and $E_{\phi}$ metrics by 4.4\% and 3.9\%, respectively, but also provides deeper insight into hidden objects, boosting the mDice and mIoU metrics by 13.2\% and 14.5\%, respectively. In summary, we have thoroughly validated the effectiveness of the different cue generation components.

\section{Conclusion}

In this paper, we propose a novel VCOD method called \textsc{\textbf{Phantom-Insight}}. We construct representations of temporal and spatial clues for camouflaged objects in videos and perform feature fusion via LLM to generate multiple cues that guide the SAM model. Additionally, we introduce a decoupled foreground-background learning strategy to enhance the separability of objects and backgrounds. Experiments on the MoCA-Mask and CAD2016 datasets demonstrate that the proposed \textsc{\textbf{PHANTOM-INSIGHT}} achieves state-of-the-art performance across various evaluation metrics.
\bibliographystyle{ACM-Reference-Format}
\bibliography{Adaptfusion}

\end{document}